\documentclass{article}
\usepackage{spconf,amsmath,graphicx}
\usepackage{times}
\usepackage{algorithm}
\usepackage{latexsym}
\usepackage{array}
\usepackage{booktabs}
\usepackage{stfloats}
\usepackage{algpseudocode}
\usepackage{float}
\usepackage{multirow}
\usepackage{adjustbox}
\usepackage{makecell}
\usepackage{caption}
\usepackage{subcaption}
\usepackage{xcolor}
\usepackage{tabularx}
\usepackage{svg}
\usepackage{enumitem}
\usepackage{hyperref}

\newcolumntype{L}{>{\raggedright\arraybackslash}X}
\newcolumntype{C}{>{\centering\arraybackslash}X}
\newcolumntype{R}{>{\raggedleft\arraybackslash}X}

\ninept
\name{
\begin{tabular}{@{}c@{}}
Min-Han Shih$^{1\ast}$, Ho-Lam Chung$^{1\ast}$, Yu-Chi Pai$^{1\ast}$, Ming-Hao Hsu$^1$, \\Guan-Ting Lin$^1$, Shang-Wen Li$\dagger$, Hung-yi Lee$^1$
\end{tabular}
\thanks{ $^\ast$ Equal Contribution.}
}

\address{$^{1}$Electrical Engineering, National Taiwan University \\ $^{\dagger}$Meta AI}
\begin{document}
\title{GSQA: An End-to-End Model for Generative Spoken Question Answering}
\maketitle
\abstract
    In recent advancements in spoken question answering (SQA), end-to-end models have made significant strides. However, previous research has primarily focused on extractive span selection. While this extractive-based approach is effective when answers are present directly within the input, it falls short in addressing abstractive questions, where answers are not directly extracted but inferred from the given information. To bridge this gap, we introduce the first end-to-end Generative Spoken Question Answering (GSQA) model that empowers the system to engage in abstractive reasoning. The challenge in training our GSQA model lies in the absence of a spoken abstractive QA dataset. We propose using text models for initialization and leveraging the extractive QA dataset to transfer knowledge from the text generative model to the spoken generative model. Experimental results indicate that our model surpasses the previous extractive model by 3\% on extractive QA datasets. Furthermore, the GSQA model has only been fine-tuned on the spoken extractive QA dataset. Despite not having seen any spoken abstractive QA data, it can still closely match the performance of the cascade model. In conclusion, our GSQA model shows the potential to generalize to a broad spectrum of questions, thus further expanding the SQA capabilities of abstractive QA.
    
\section{Introduction}
Question Answering (QA) tasks have consistently emerged as one of the foundational challenges~\cite{DBLP:conf/iclr/LewisF19,DBLP:conf/emnlp/ShakeriSZNNWNX20,DBLP:conf/emnlp/YueKF21,DBLP:conf/coling/YueZKS022}. As a touchstone for measuring machine comprehension of textual content, QA presents unique complexities that have long captivated researchers. If answers are directly in the text and just need to be found, we categorize it as extractive QA. In contrast, when the answers are not explicitly present, needing a deeper dive involving inference and synthesis to craft a response, we refer it to as abstractive QA. With the rise of smart speakers and voice assistants, an emerging challenge in QA is Spoken Question Answering (SQA) to facilitate interactions between humans and these assistants. SQA requires both the questions and answers to be vocalized. Unlike text, speech encompasses richer details such as the identity of the speaker, their tone, and more, which introduces added intricacies to the SQA tasks.

An intuitive approach to tackle this challenge is to cascade models. Here, an Automatic Speech Recognition (ASR) system first transcribes the input speech of both questions and passages and then feeds transcriptions into a textual question answering language model that predicts the corresponding answers. Last, answers are transformed back into speech through a Text-To-Speech (TTS) model. The cascaded method's primary advantage is its ability to harness the great power of text-based language models trained on vast textual datasets, ensuring it can tackle both extractive and abstractive QA tasks. However, the cascaded model has its drawbacks. The primary issue stems from error propagations, since the text-based language models are not typically trained on ASR-erroneous data, the presence of ASR errors can notably undermine the performance of cascaded models. Moreover, some natural languages do not possess writing text, which leads to the absence of language models that are capable of acquiring these languages. Thus, to tackle these issues, an End-to-End speech model is needed.

\begin{figure}[t]
    \begin{adjustbox}{max width=\linewidth}
      \centering
      \includegraphics[]{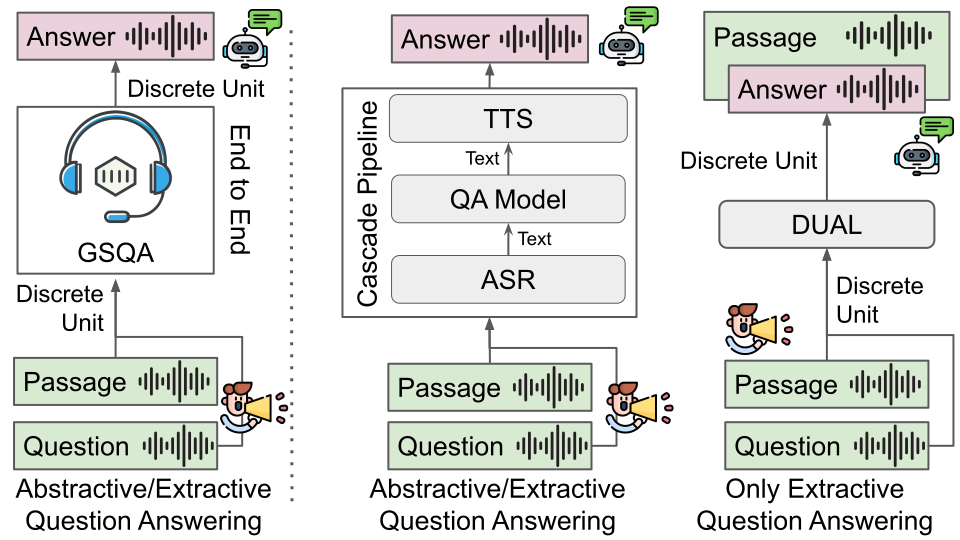}
    \end{adjustbox}
    \caption{GSQA compared to other baselines: The Cascade Method accommodates both abstractive and extractive QA but risks error propagation. DUAL is an end-to-end textless approach, exclusive to extractive QA. GSQA is a textless, end-to-end generative method, capable of handling both extractive and abstractive QA.}
    \label{fig:overview}
\end{figure}

Recently, end-to-end, textless approaches (e.g., DUAL\cite{DUAL}) have emerged as a solution to the cascade method's limitations. These works \cite{DUAL, discretize, hubertunit} rely on the encoded representations of acoustic models, like HuBERT\cite{hubert}, and apply K-means clustering to transform speech representations into discrete units. These units, encapsulating both questions and passages, are fed into the model that predicts the span of the answer in the passage using start and end positions. By adopting this textless approach, DUAL effectively sidesteps the error propagation from ASR systems~\cite{DBLP:conf/interspeech/LeeWLL18,DBLP:conf/icassp/LeeCL19,DBLP:conf/icassp/YouCZ21}. It improves the model robustness in noisy speech. However, there is a notable limitation to the DUAL model: it is not able to handle abstractive QA scenarios, as it is designed to only output start and end positions within the input spoken passage.

As Fig.\ref{fig:overview} shows the difference between our proposed method, cascaded method, and DUAL, we combine the advantages of both the cascaded method and end-to-end (E2E) textless approach to introduce the first-ever textless E2E model capable of handling both extractive and abstractive QA. Mirroring DUAL's methodology, we utilize HuBERT to convert the input speech (questions and passages) into discrete units\cite{discretize}. A sequence-to-sequence model then processes these units to generate answers in the form of discrete units. Motivated by the strengths and limitations of the approaches above, we propose GSQA (Generative Spoken Question Answering), the first textless end-to-end model capable of handling both extractive and abstractive QA.

Furthermore, given the current absence of a spoken abstractive QA corpus, we begin by initializing GSQA with a generative Text QA (TQA) model. This TQA model has been trained on several extractive and abstractive datasets using a generative approach. Subsequently, we fine-tune the TQA model on NMSQA, a dataset dedicated to extractive spoken QA. The primary objective of GSQA is to impart the Natural Language Understanding prowess of the TQA model to GSQA. We also use Microsoft's Azure TTS service to create a synthesized test set for evaluation. It's noteworthy that, during the transfer learning phase, GSQA exhibited a significant enhancement, registering an 18\% relative gain in the BLEU1 score in a zero-shot abstractive spoken QA setting.\\
The contributions of this paper are summarized below:\setlist{nolistsep}
\begin{enumerate}[noitemsep]
    \item The introduction of spoken generative question answering model GSQA, the first end-to-end textless generative model. 
    \item The establishment of a method where pre-training on textual QA, followed by fine-tuning on extractive spoken QA, leads to zero-shot abstractive spoken QA.
    \item Demonstrating that our proposed model has competitive performance against textless extractive models.
\end{enumerate}
\section{Related Work}
\label{sec:related}
The existing SQA methods primarily utilize Automatic Speech Recognition (ASR)~\cite{wav2vec2, whisper} transcripts, processed by a large language model (LLM) for QA tasks. A key challenge is that ASR errors significantly impact the LLM's performance. For instance, less robust ASR systems produce incorrect transcripts, leading to erroneous LLM predictions. In contrast, more advanced ASR systems, such as Whisper-large with 1550M parameters, offer lower error rates and provide more accurate input for the LLM, improving QA task accuracy. However, these high-parameter models also result in longer inference times. In extractive SQA, models identify answer spans within spoken documents. DUAL~\cite{DUAL} is noteworthy for its ability to determine these spans, but it fails when answers are not present in the input audio. TWIST~\cite{twist}  addresses SpeechLM's~\cite{speechlm} limitations, particularly its lack of semantic understanding, which often results in irrelevant or incorrect responses. While TWIST improves SpeechLM's initialization, it still shows limited progress in transferring semantic knowledge from text to speech.

\section{Methodology}
\vspace{-1em}
\subsection{Overview \& Formulation}
In the realm of Question Answering (QA), we divide the problem into three components: the Question (\textit{Q}), the Passage (\textit{P}), and the Answer (\textit{A}). These entities can manifest in two formats: textual data (\(Q_t\), \(P_t\), \(A_t\)) and speech discrete unit-based data (\(Q_u\), \(P_u\), \(A_u\)).

\begin{figure}[t]
    \begin{adjustbox}{max width=\linewidth}
      \centering
      \includegraphics[]{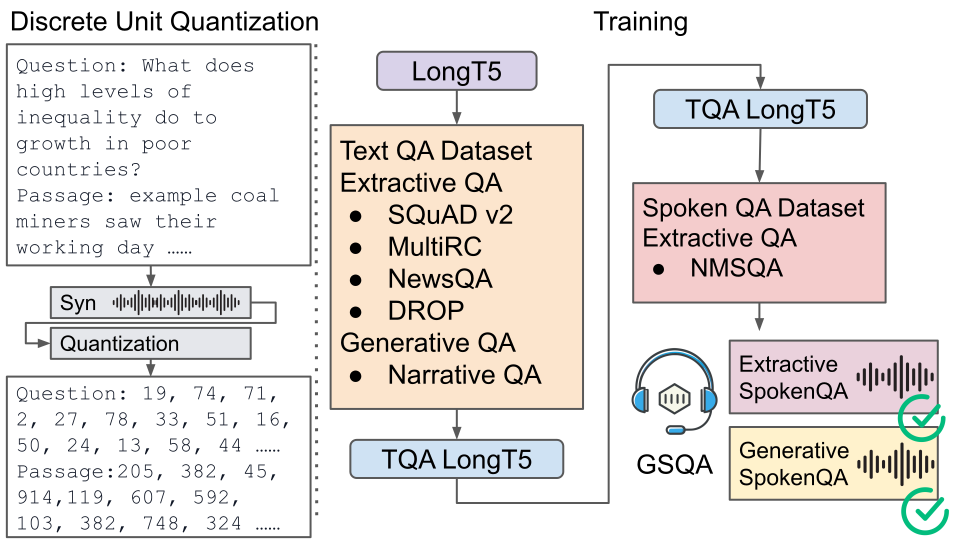}
    \end{adjustbox}
    \caption{\textbf{Left:} The process of discrete unit quantization from synthesis data.  \textbf{Right:} Model Training Procedure: A depiction of the transition from textual QA pretraining to spoken QA fine-tuning.}
    \label{fig:training-procedure}
\end{figure}

Our method lies in the ambition to create a textless generative end-to-end language model that is capable of question answering. Initially, we follow \cite{discretize, hubertunit} to quantize speech representations into discrete units. We then utilize a discrete unit-based sequence generative model to craft the answer units, which would be subsequently transformed into speech via a Unit-based HiFi-GAN vocoder~\cite{hifigan}. To boost the logical understanding capabilities of the discrete unit-based sequence generative model, we initialize this generative SQA model using weights from a textual model~\cite{twist}. Fig.\ref{fig:training-procedure} shows the entire model architecture and training pipeline.

\vspace{-1em}
\subsection{Speech Quantization}
HuBERT~\cite{hubert} is used for efficient speech representation, converting speech signals into discrete units through mask segment prediction training. This involves clustering each 20ms frame into one of K categories using k-means clustering, where the centroid ID of each cluster represents the frame's discrete unit. To achieve this conversion, we utilize the HuBERT-base model with a 100-cluster configuration, following Lee \textit{et al.}'s methods~\cite{discretize}. Additionally, we employ run-length encoding to condense consecutive identical units into a single unit, thus reducing the length of input unit sequences.
\vspace{-1em}
\subsection{Text Question Answering Pretraining}\label{tqa}
Due to the scarcity of training data in abstractive spoken QA, our model's performance could be limited. To address this, we propose using a pretrained textual model to enhance semantic understanding. This model is initially trained as a generative language model and then adapted for SQA, facilitating the transfer of textual QA capabilities to speech units.

Handling the longer sequences in speech, even after deduplication, presents a challenge, as many generative textual models struggle with such long inputs~\cite{bart}. Therefore, our chosen initialized model must be generative and capable of processing sequences longer than 4096 tokens~\cite{llama, longt5}.

We select LongT5~\cite{longt5} for this purpose. LongT5, an extension of the T5 encoder, incorporates global-local attention mechanisms to efficiently manage longer inputs. It combines the attention efficiency of ETC with PEGASUS's summarization-focused pre-training, offering improved performance. We further pretrain LongT5 on textual QA datasets, denoted as \textbf{LongT5-TQA}.

The primary goal during this phase is to train the model to process a textual question \(Q_t\) and passage \(P_t\) and generate the corresponding answer \(A_t\). This pretraining is crucial, equipping our model with foundational QA skills for the subsequent phases of our training methodology.

\subsection{GSQA: Generative Spoken QA Model}
Fig.\ref{fig:training-procedure} illustrates our GSQA training procedure. As mentioned in Sec.\ref{tqa}, we pretrain the LongT5 on Text Question Answering (TQA) tasks to produce LongT5-TQA, which is capable of text-level extractive and abstractive QA. This model is then fine-tuned on the Extractive Spoken QA dataset, NMSQA~\cite{DUAL}. To unify the dataset format in abstractive QA form, we modify the labels in NMSQA from time spans to the answers' waveform. Here, the given spoken passages, questions, and answers would go through a quantization model and would be converted into discrete units. Since LongT5-TQA does not recognize discrete units, we introduced them as new tokens. The embeddings of units are initialized by random sampling from other text token embeddings, ensuring proper tokenization and a seamless transition from text to speech. By these modifications, LongT5-TQA would be fine-tuned to predict answers in discrete units instead of text, where we denote the fine-tuned model as \textbf{Unit-LongT5-TQA}. Note that if the model's denotation is not appended with TQA, it means that we don't pretrain the model on TQA tasks. Last, to achieve speech-to-speech, we employ a HiFi-GAN vocoder~\cite{lee-etal-2022-direct} to convert discrete units back to speech. This vocoder includes a duration prediction module for deduplicated unit synthesis. 
\begin{table*}
\centering
\begin{adjustbox}{max width=\linewidth}
   \begin{tabular}{lc|cc|cc|cc}
    \toprule
    \multirow{2}{*}{} & \multicolumn{1}{c}{} &\multicolumn{2}{c}{\textbf{NMSQA}-dev} & \multicolumn{2}{c}{\textbf{NMSQA}-test} & \multicolumn{2}{c}{\textbf{Spoken-NarrativeQA}-test} \\
     \multicolumn{1}{c}{Model} & Num. Param. & F1 Score & EM Score & F1 Score & EM Score & BLEU1 & ROUGE-L \\ 
    \midrule
    Cascade model (w/ ASR transcriptions) & 1025M & 49.1 & 32.0 & 47.3 & 30.4 & 13.5 & 19.9 \\ \midrule
    DUAL & 452M & 39.4 & 21.9 & 33.6 & 21.2 & - & - \\
    Unit-LongT5 & 312M & 25.5 & 12.6 & 20.1 & 9.4 & 6.8 & 10.4 \\
    Unit-LongT5-TQA (Proposed) & 312M & \textbf{41.8} & \textbf{24.9} & \textbf{36.0} & \textbf{24.0} & \textbf{8.0} & \textbf{11.8}\\
    \bottomrule
    \end{tabular}
\end{adjustbox}
\caption{The table presents the speech-to-speech evaluation of our models and the number of parameters for each method. For extractive QA datasets, we utilized F1 and exact match (EM) as evaluation metrics. On the other hand, for abstractive QA datasets, we employed BLEU1\cite{bleu} and Rouge-L\cite{rouge} for assessment.  The cascaded model in this table includes an ASR system and an LM, which is \textbf{Whisper-medium.en} \cite{whisper} and \textbf{LongT5-TQA} \cite{longt5}, respectively. DUAL\cite{DUAL} is combined a \textbf{HuBERT-base} \cite{hubert}, \textbf{K-means model} for 128 clusters, and a \textbf{Longformer} \cite{longformer}.}
\label{tab:experiment}
\end{table*}
\vspace{-1em}
\section{Experimental Setup}
\subsection{Dataset}
\label{dataset}
For TQA pretraining data, We utilized four extractive QA datasets: SQuAD~\cite{squadv2}, MultiRC~\cite{superglue}, NewsQA~\cite{newsqa}, and Drop~\cite{drop}, along with one abstractive dataset, NarrativeQA~\cite{narrativeqa}, to pretrain LongT5-TQA, and use it as foundation model of GSQA. On the other hand, for SQA training, we use NMSQA~\cite{DUAL} as our downstream dataset for training Unit-LongT5. The NMSQA is derived from SQuAD v1.1~\cite{squadv1} and consists of Wikipedia paragraphs with human-written questions. While its train and dev sets feature synthesized speech using Amazon Polly, the test set contains audio from 60 human speakers (30 males and 30 females). In addition, we propose a new spoken abstractive SQA evaluation set, named \textbf{Spoken-NarrativeQA}. This dataset is built upon the NarrativeQA~\cite{narrativeqa} test set, an abstractive text question answering dataset for reading comprehension from books and movie scripts with complex narrative understanding. Note that many answers in this dataset are not directly extractable from passages. We transform the text to speech using the TTS system and specifically select test data where answers are absent in the passage. There are 1626 testing samples in the Spoken-NarrativeQA dataset. 

\begin{table}[t]
    \centering
    \begin{adjustbox}{max width=\linewidth}
    \begin{tabular}{lccc}
    \toprule
         \textbf{Method} & \textbf{Speech-to-(text/unit) model} & \textbf{QA Model} & \textbf{Total} \\
         \midrule
         Cascade model & Whisper-medium.en (764M) & LongT5-TQA (261M)&  1025M \\
         DUAL & HuBERT-large-128-layer$_{1-22}$ (304M) & LongFormer (148M) &  452M \\
         Unit-LongT5-TQA & HuBERT-base-100-layer$_{1-6}$ (51M) & LongT5-TQA (261M) &  312M \\ \bottomrule
    \end{tabular}
    \end{adjustbox}
    \caption{The respective number of parameters in our experiment. Note that the E2E method, like DUAL and our work, does not need an ASR system but needs a Unit-Extraction model to generate units, and the number after the name of that is the number of clusters.}
    \label{tab:param}
\end{table}
\begin{table}[t]
\centering
   \begin{adjustbox}{max width=\linewidth}
   \begin{tabular}{ccc}
   \toprule
  \textbf{NMSQA}-dev & \textbf{NMSQA}-test  & \textbf{Spoken-NarrativeQA}-test \\ \midrule
       8.1 & 14.8  &  6.3  \\
    \bottomrule
    \end{tabular}
    \end{adjustbox}
\caption{Word Error Rate (WER) of the pretrained Whisper-medium.en ASR system on different datasets for Spoken QA tasks.}
\label{tab:asr}
\end{table}

\begin{table}[h!]
  \centering
  \begin{adjustbox}{max width=\linewidth}
  \begin{tabular}{|l|l|}
    \hline
    \textbf{Question} & What did Tancred's destiny seem to be? \\
    \hline
    \textbf{Answer} & To live the life of a normal member of the British ruling class. \\
    \hline
    \textbf{Unit-LongT5} & live the life of any \\
    \hline
    \textbf{Unit-LongT5-TQA} & live the life of any conventional member of the British ruling class \\
    \hline
  \end{tabular}
  \end{adjustbox}
  \caption{Comparison of Results}
  \label{tab:results}
\end{table}
\subsection{Cascaded Pipeline}
The conventional approach for SQA tasks typically follows a cascaded pipeline, as illustrated in the middle of Fig.~\ref{fig:overview}. The passages and questions are first transcribed by an ASR system, followed by a text-based language model that generates text-based answers, and a TTS system converts the text-based answers into speech answers. To be more specific, we use \texttt{Whisper-medium.en}~\cite{whisper} as the ASR system, the fine-tuned LongT5-TQA model as the text-based language model, and the Azure TTS service as the TTS system. We record the cascaded pipeline's experiment result in the Cascade Model column in Table~\ref{tab:experiment}. The total number of parameters of the cascaded model is shown in Table~\ref{tab:param}, and the WERs of the ASR system in given datasets are listed in Table~\ref{tab:asr}.
\subsection{Automatic Performance Evaluation} 
Since all our models ultimately output in speech, what we focus on is whether the content of this waveform contains the answer. We then use an ASR model to convert the generated waveform back to text, allowing us to use text-based metrics to evaluate the content of the model. To reduce errors from the ASR during evaluation, we utilize the state-of-the-art ASR model, \texttt{whisper-large-v2}, during automatic performance evaluation. We then compare the transcriptions of the outputs from \texttt{whisper-large-v2} with the text ground truth. We calculate the F1-score and Exact-matched (EM) scores for extractive QA tasks. The BLEU1 score \cite{bleu} and the Rouge score \cite{rouge} are used for abstractive QA tasks. Lastly, the experiment results of all models are reported in Table~\ref{tab:experiment}.

\subsection{Implementation details}
As the previous section mentioned, we pretrain LongT5 on the 5 TQA datasets for 13 epochs, the pretraining learning rate is set to 0.0005, and the weight decay is 0.01. On the other hand, we fine-tune the pretrained LongT5 on the downstream unit-based dataset, NMSQA, for 25 epochs, the downstream learning rate is set to 0.0003, and the weight decay is 0.001. Last, when we inference our model on the abstractive test set(e.g., \textbf{Spoken-NarrativeQA}-test), to encourage GSQA output longer answers, we use beam search for decoding with beam size of 5, and set length penalty to 2. 

\section{Results}
As illustrated in Table~\ref{tab:experiment}, we evaluate our models on both extractive QA dataset NMSQA-dev, NMSQA-test, and on abstractive QA dataset Spoken-Narrative QA. Among the models evaluated, the proposed \textbf{Unit-LongT5-TQA} exhibits remarkable performance. On the \textbf{NMSQA}-dev set, it achieves an F1 score of 41.8\% and an EM score of 24.9\%. Its performance remains consistent on the \textbf{NMSQA}-test dataset, registering scores of 36.0\% and 24.0\% for F1 and EM, respectively. Furthermore, this model outperforms the others on the generative \textbf{Spoken-NarrativeQA}-test set, securing a BLEU1 score of 8.0\% and a ROUGE-L score of 11.8\%.
\subsection{The effectiveness of TQA pretraining}
From Table\ref{tab:experiment}, we can observe that Unit-LongT5-TQA has huge improvements on both NMSQA-dev and NMSQA-test, which gets 16\% F1-scores gain than Unit-LongT5 and gets two times than it on EM scores. It shows that pretraining language model on TQA tasks brings significant advantages for unit-based model initialization to learn semantic information in speech. On the other hand, in the Spoken-NarrativeQA-test, an abstractive spoken QA dataset, Unit-LongT5-TQA also outperforms Unit-LongT5 on both BLEU1 and ROUGE-L scores, which gets 1.2 and 1.4 increase, respectively. In our showcase, we have observed that TQA pretraining could further help the model generating more fluent sentences. The following results in Table~\ref{tab:results} are from our test set. Without TQA pretraining, our prediction is "live the life of any," where the sentence is truncated in the middle, resulting in an incoherent meaning. However, after TQA training, the model would predict complete sentences.

\subsection{Performance at different WERs}
\begin{figure}[t]
    \centering
    \includegraphics[width=0.87\linewidth]{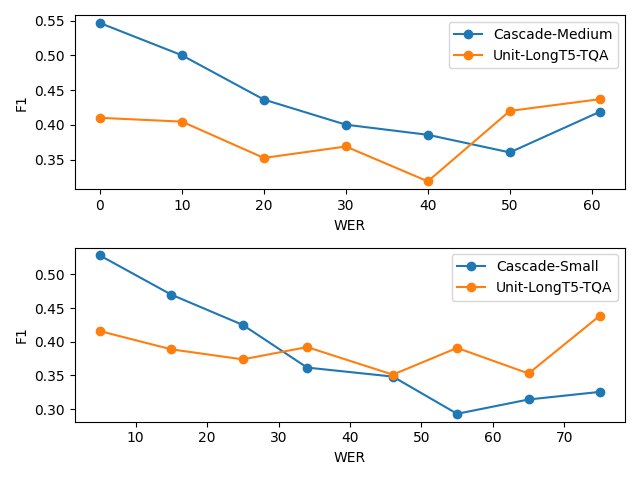}
    \caption{We sample the 8000 data within NMSQA-dev to verify the impact on the cascaded model under different Word Error Rates (WERs) with different ASR systems.}
    \label{fig:wer}
\end{figure}

Fig.~\ref{fig:wer} illustrates the superior stability and efficiency of our E2E model for SQA, especially compared to the cascaded model's sensitivity to ASR WERs. The cascaded model, using ASR systems \textbf{Whisper-medium.en} and \textbf{Whisper-small.en}, demonstrates marked performance degradation with increasing WERs, whereas our E2E model maintains consistency even in high WER scenarios. Additionally, the cascaded model, particularly with Whisper-medium.en, requires significantly larger parameters, escalating computational resource demands. While enhancing the ASR system improves cascaded model performance, this approach requires extensive paired text-speech data, larger model scales, and considerable training costs for both LLMs and the TTS system, making it extremely expensive. In contrast, our smaller E2E model potentially offers comparable performance with greater training efficiency. Furthermore, the E2E model circumvents the limitations of languages without written forms, which are insurmountable for ASR-based systems. These advantages firmly position our E2E model as a more stable, resource-efficient, and universally applicable solution for SQA tasks.

\subsection{Models parameters}
For a fair comparison, we strive to ensure the number of parameters across all models remains relatively consistent. Upon examination, the cascaded method is found to have a large number of parameters. This is primarily because it requires stacking three huge models. Since the performance of each model should not be compromised, opting for smaller models is not feasible, leading to a larger overall parameter size. In contrast, end-to-end models rely on a single model to accomplish SQA tasks, resulting in a reduced number of parameters. We observed that for abstractive QA, there isn't a stringent requirement for a compelling speech-to-unit model. It is sufficient to encode the content's information into discrete units. Instead, a more potent QA model is needed to infer the answers to the questions. Detailed parameter counts are presented in Table~\ref{tab:param}. Our model has the fewest parameters among all models. Notably, our model outperforms the others across various metrics.
\section{Conclusion}
In conclusion, we introduce the first end-to-end Generative Spoken Question Answering (GSQA) model, bridging the gap between spoken extractive and abstractive QA tasks. Combining the advantages of cascaded and textless models, our GSQA model demonstrates superior performance on extractive QA datasets, outperforming the previous E2E extractive SQA models. Furthermore, in the challenging abstractive zero-shot domain, our model exhibits competitive capabilities. We also highlighted the importance of pretraining the language model on textual data to enhance the unit-to-unit model's capability of semantic understanding. Our future research would concentrate on improving the model's performance with human narrations and exploring training strategies to enhance generalization ability across diverse SQA tasks.
\clearpage
\section{Acknowledgement}
We thank the National Center for High-performance Computing (NCHC) of National Applied Research Laboratories (NARLabs) in Taiwan for providing computational and storage resources.
\bibliographystyle{IEEEbib}
\bibliography{refs}

\end{document}